# Generative AI for Banks: Benchmarks and Algorithms for Synthetic Financial Transaction Data

*Submission Type: Completed Full Research Paper*

Fabian Sven Karst
University of St.Gallen
fabian.karst@unisg.ch

Sook-Yee Chong
Aboitiz Data Innovation
sookyee.chong@aboitiz.com

Abigail A. Antenor
Aboitiz Data Innovation
abigail.antenor@aboitiz.com

Enyu Lin
Aboitiz Data Innovation
enyu.lin@aboitiz.com

Mahei Manhai Li
University of St.Gallen; University of Kassel
mahei.li@unisg.ch

Jan Marco Leimeister
University of St.Gallen; University of Kassel
janmarco.leimeister@unisg.ch

## Introduction

The banking sector, as a data-driven industry, relies on the availability of high-quality data to create value and protect its customers. The synergy between recent deep learning (DL) advancements, and the sector's data needs presents a growth potential of USD$4.6 trillion by 2035 (Accenture, 2017). However, deploying DL models is challenging due to the need for large, high-quality training data (Ryll et al., 2020), a difficulty made worse by the intricacy of financial transaction data (with complex data patterns and time-related characteristics), and strict regulations that limit data sharing (EU Regulation 2016/679, PCI DSS v4.0). One possible solution is to use synthetic data which is artificially generated rather than drawn from real-world events to increase samples in the minority class (Jordon et al., 2022), and allow safe data sharing between financial institutions while protecting privacy (Karst et al., 2024). This approach is essential for improving models used in assessing risks and detecting fraud.

This potential has led to numerous studies proposing algorithms for generating financial trans-action data (Strelcenia and Prakoonwit, 2023; Pandey et al. 2020; Sethia et al. 2018). However, these studies have notable limitations: most algorithms (74.29%) are tested on the small European Credit Cardholder dataset, which includes only 284,807 transactions, and is anonymized, hindering analysis of key features like its graph structure. Additionally, inconsistent performance metrics make algorithm comparisons difficult.

Thus, this paper addresses the following research questions (RQ). **RQ1:** What methods and evaluation criteria exist for generating synthetic financial transaction data? **RQ2:** Which algorithm is most effective for generating synthetic financial transaction data?

## Methods

### *Systematic Literature Review*

To develop algorithms for replicating financial transactions, we reviewed the literature on synthetic data generation techniques specific to the financial domain. This review followed the PRISMA methodology (Page et al. 2020). Based on RQ1, the following search string was formulated:

*("Synthetic Data" OR "Artificial Data" OR "Simulated Data" OR "Generated Data" OR "Mock Data" OR "Data Synthesis" OR "Data Augmentation") AND ("Payment*" OR ("Bank*" AND "Transaction") OR ("Financ*" AND "Transaction") OR ("Sequen*" AND "Bank*") OR ("Sequen*" AND "Transaction") OR ("Sequen*" AND "Financ*") OR ("Financ*" AND "Network") OR ("Bank*" AND "Network") OR ("Transaction" AND "Network"))*

Using this string, we searched the four prominent scientific databases - Web of Science, Scopus, IEEE Xplore, and ACM Digital Library - for English-language, peer-reviewed papers from the last 10 years (2013-2023), resulting in 1,216 papers. We then applied the PRISMA methodology for selection, involving initial screening and review by at least two reviewers (Figure 1). For filtering the quality criteria from Carrera-Rivera et al. (2022) and Dybå and Dingsøyr (2008) were applied (Figure 1). Ultimately, 30 papers met the criteria for inclusion in the review.

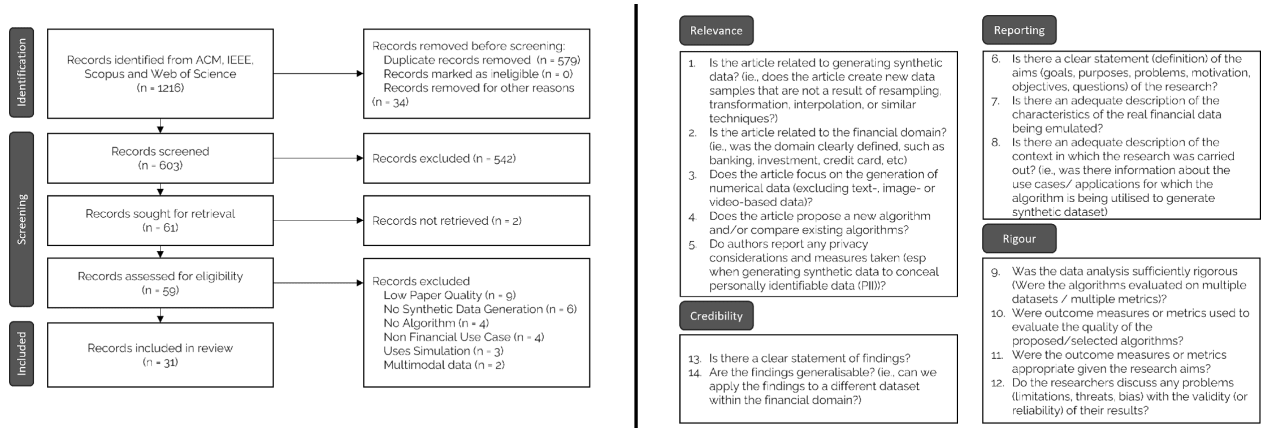

**Figure 1: PRISMA Flow Diagram (left) and Assessment Criteria with Questions (right)**

*Algorithm Evaluation*

The identified algorithms were implemented in Python (GitHub) and optimized on a 100,000-row subset. Final models were trained on the full dataset and evaluated against the training data. To ensure validity, this process was applied to two financial transaction datasets (one real, one simulated). The complete real-world dataset consists of anonymized data covering a one-month period and amounting to 5,202,003 transactions. Although the specific details of its features cannot be disclosed due to confidentiality constraints, they encompass information related to the transaction, the source and target tokens, the amount, and transfer details. After preprocessing and removing rows with any missing values 5,201,946 transactions remained. The second dataset, simulated by IBM, contains 4,211,370 rows, is publicly available, and shares similar characteristics with the real dataset. For both datasets, graph embeddings, based on Abu-El-Haija et al. (2018), were generated to capture structural relationships within the network.

## I – Algorithm and Benchmark Identification

This section highlights the algorithms and the performance metrics identified from the systematic review within the financial domain.

### *Algorithms*

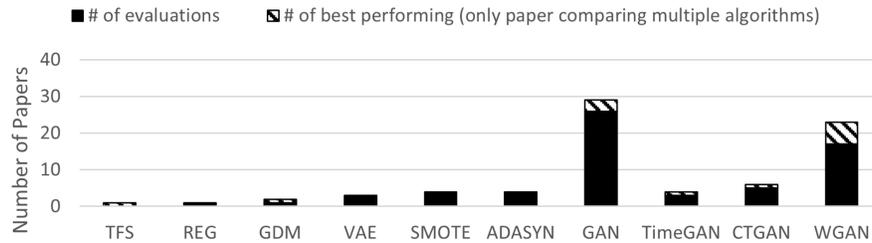

**Figure 2: Number of evaluations per algorithm and number of times each category performed best**

Generative Adversarial Networks (GANs), first introduced by Goodfellow et al. (2014), are the dominant choice for generating synthetic financial transaction data, comprising 70% of proposed architectures (Figure 2). However, recent papers (post-2023) show that only 50% of top-performing architectures are GAN-based.

### GAN-based architectures

**Wasserstein GAN (WGAN)** – WGAN improves GAN architecture, by using Wasserstein distance and methods like weight clipping or gradient penalty for training stability (Arjovsky et al. 2017; Gulrajani et al. 2017; Mescheder et al. 2018), with gradient penalty being the most common (62% of WGANs). Modifications have been made to address the temporal nature of transaction data (Bratu and Czibula, 2021; Zhang et al. 2020; Li et al. 2020; Strelcenia and Prakoonwit, 2023). Pandey et al. (2020) showed that using the discriminator model for sample quality validation enhances WGAN's performance. WGAN is particularly effective when training data is limited or during oversampling of minority classes (Gangwar and Ravi, 2019; Hilal et al. 2022; Sethia et al. 2018; Strelcenia and Prakoonwit, 2023).

**Conditional Tabular GAN (CTGAN)** – CTGAN enhances WGAN with mode-specific normalization and conditional generator sampling, allowing it to synthesize data with continuous columns of arbitrary distributions and handle highly imbalanced data (Xu et al. 2019) and was able to outperform other models in the study by Hong and Baik (2021).

**Other GAN-based methods** – Additionally, 23 different GAN adaptations were evaluated in the analyzed papers. We focus on DoppelGANger (DGAN), which outperformed the other algorithms in its respective studies. DGAN (Lin et al. 2020) improved GAN architecture by separating time series and metadata generation, normalizing data, and using batch generation to boost temporal correlation, leading to better performance in stock market data generation (Dogariu et al. 2021).

**Non-GAN-based architectures**

Non-GAN architectures fall into two categories: traditional oversampling methods, commonly used as baselines, and new deep learning-based methods, which will be the focus of this section.

**Variational AutoEncoders (VAE)** - VAEs train two neural networks to compress and decompress data into a low-dimensional space, with the decompression module used to generate new synthetic data from random inputs (Kingma and Welling, 2022). Most papers analyzed use a modified version, Tabular Variational Autoencoder (TVAE), which employs mode-specific normalization and conditional generator sampling to generate continuous data with arbitrary distributions and handle highly imbalanced data. While less frequently used, papers that employ VAEs consistently rank among the top algorithms (Hong and Baik, 2021; Sattarov et al. 2023).

**Gaussian Diffusion Models (GDM)** - Building on the denoising diffusion probabilistic model (Ho et al. 2020) and extending it with an embedding layer, GDMs introduce a novel architecture for generating mixed-type financial tabular data (Sattarov et al. 2023). Sattarov et al. (2023) demonstrate that their Financial Diffusion (FinDiff) model, a type of GDM, effectively captures both individual and joint distributions while preserving data privacy, outperforming TVAE and CTGAN across various metrics.

*Benchmarks*

Evaluating synthetic data quality is essential for understanding its usefulness. However, this is challenging due to the wide range of metrics employed, and the lack of consensus on the most effective ones (Dogariu et al. 2021). Based on the metrics most frequently cited in the reviewed articles, we developed an evaluation framework adapted from Sattarov et al. (2023). This framework consists of five categories: fidelity, synthesis, efficiency, privacy, and graph structure.

We added graph structure as a criterion due to its importance in analyzing transaction flows and detecting fraud in financial institutions.

**Fidelity -** This is a measure of how closely the synthetic data resembles the original data. Reviewed papers often used visualizations like pair plots, PCA, and t-SNE to compare distribution shapes, central tendencies, and variability (Bratu and Czibula, 2021). However, visual interpretations can be subjective. Quantitative measures such as Kullback-Leibler divergence, Jensen-Shannon divergence, Earth Mover's Distance, and Kolmogorov-Smirnov (KS) statistic provide objective comparisons (Dogariu et al. 2021; Ferreira et al. 2021; Hong and Baik, 2021; Li et al. 2020; Sattarov et al. 2023). For bivariate comparisons, Spearman or Pearson correlations are used (Ferreira et al. 2021). In this study, we conduct two types of analysis: column-wise fidelity and row-wise fidelity. Column-wise fidelity assesses how well each column of synthetic data replicates the corresponding column in the real data, using the KS statistic, where a higher value indicates greater similarity. Row-wise fidelity evaluates the correlation between pairs of columns using Pearson correlation, with a higher score indicating higher similarity between the datasets.

**Synthesis** - The synthesis metrics evaluate a model's capability to generate synthetic data that is not an exact replication of the original data. Several reviewed articles used divergence scores, such as Kullback-Leibler, Jensen-Shannon, and Earth Mover's Distance, to optimize GANs and closely match synthetic data distributions to original data (Dogariu et al. 2021). Although not direct comparison metrics, these scores measure the discrepancy between distributions, making them useful benchmarks for evaluating data synthesis. We adopted the approach from Sattarov et al. (2023) to assess whether the generated records are novel or exactly replicate records from the original dataset within a 1% margin.

**Computational efficiency -** This metric measures the time (in seconds) to train the algorithms, balancing model complexity and practicality. High efficiency indicates the algorithm can quickly process large datasets with minimal resources. It is calculated by averaging the training time across 30 runs with different hyperparameters, using 100,000 samples to generate 10,000 new synthetic records.

**Privacy -** The privacy metric indicates how effectively the synthetic data prevents the identification of the original data entries. Only two reviewed papers reported privacy measures. Park et al. (2021) used 'disclosure risk' to assess the likelihood of re-identifying personal information in consumer credit data. Sattarov et al. (2023) employed the Distance to Closest Records (DCR) metric, measuring the median Euclidean distance between synthetic and original data entries. We also used the Nearest Neighbour Distance Ratio (NNDR), which compares the Euclidean distances to the closest and second closest neighbors. To provide a robust estimate of privacy risk, we calculated the 5th percentile of both metrics (Zhao et al. 2021). Higher DCR or NNDR scores indicate a lower risk of synthetic data revealing real information.

**Graph Structure -** We conducted a comparison between the graph structures of the real and synthetic data, using the NetSimile score as outlined by Berlingerio et al. (2012). A higher score indicates a greater similarity between the networks.

## II - Algorithm Performance Evaluation

This section compares the identified algorithms. Full details on hyperparameter search and analysis are available on GitHub.

| | column-wise Fidelity score | row-wise Fidelity score | Synthesis score | Privacy - DCR 5th percentile | Privacy - NNDR 5th percentile | Efficiency | Graph Structure-NetSimile |
|---|---|---|---|---|---|---|---|
| CTGAN | 0.87787 | 0.95258 | 0.99998 | 0.01968 | 0.98550 | 2267 s | 30.96229 |
| DGAN | 0.47808 | 0.86198 | **1.0** | **1.81121** | 0.94122 | 743 s | **30.77869** |
| WGAN | 0.24064 | 0.94509 | **1.0** | 1.77556 | **0.99992** | 538 s | NaN |
| FinDiff | **0.95429** | **0.98524** | 0.84101 | 0.00033 | 0.98781 | 625 s | 30.93536 |
| TVAE | 0.89677 | 0.97374 | 0.99922 | 0.00997 | 0.97991 | **401 s** | 31.20362 |

**Table 1: Comparative analysis of the algorithms on the real dataset**

| | column-wise Fidelity score | row-wise Fidelity score | Synthesis score | Privacy - DCR 5th percentile | Privacy - NNDR 5th percentile | Efficiency | Graph Structure-NetSimile |
|---|---|---|---|---|---|---|---|
| CTGAN | 0.35892 | 0.77140 | 0.70323 | 0.0 | 0.03617 | 6697 s | 30.38868 |
| DGAN | 0.21086 | 0.53020 | **1.0** | 0.14922 | 0.99999 | 6167 s | **29.51902** |
| WGAN | 0.10636 | 0.30918 | 0.99999 | **0.58109** | **1.0** | 1363 s | 30.47143 |
| FinDiff | **0.43746** | **0.95798** | 0.24292 | 0.0 | 0.05537 | **947 s** | 30.64564 |
| TVAE | 0.42274 | 0.90222 | 0.51400 | 0.0 | 0.06355 | 1000 s | 31.08549 |

**Table 2: Comparative analysis of the algorithms on the simulated dataset from IBM**

FinDiff achieved the highest fidelity scores (Tables 1 and 2), surpassing TVAE by 6.41% in column fidelity and 1.18% in row fidelity, in line with Sattarov et al. (2023), demonstrating FinDiff's ability to preserve the original dataset's statistical distribution. However, DGAN and WGAN excelled in data novelty, surpassing FinDiff by 15.9% in synthesis scores, indicating their synthetic data was more distinct from the original. GANs, particularly DGAN, performed well in synthesis, as expected, by promoting diversity and avoiding exact replication, while TVAE introduced variability by sampling from the original data's latent space. However, DGAN and WGAN struggled with retaining data characteristics, resulting in low column fidelity scores. CTGAN, however, demonstrates a balance between fidelity and synthesis.

The models yielded comparable graph similarity scores, except for WGAN. Specifically, the WGAN-generated graph formed a single cluster, resulting in a null value (Table 1). While WGAN does not always produce a single-cluster structure, its stability and gradient flow can lead to a more consistent dataset, which manifests as a single cluster in this study. This was observed only in the real data. The comparison of graph similarity scores within samples (real-world: 8.026, simulated data: 2.269) and between the two training datasets (25.820) indicates that the algorithms produce synthetic data that poorly resembles the original graph structure.

The efficiency metric, demonstrated by training times across different parameter combinations, showed that GAN-based architectures such as CTGAN and DGAN require significantly more training time (an increase of 65.3%). As a result, these models may not be practical for scenarios involving large datasets and limited computational resources.

Given the focus on generating synthetic financial data that includes sensitive information, privacy is a critical metric. DGAN achieved the highest DCR score (1.811) for the real dataset, while WGAN had the highest DCR score (0.581) for the IBM simulated dataset. This suggests that DGAN and WGAN may offer better privacy protection by minimizing the likelihood of reconstructing individual records from synthetic data. However, despite preserving privacy, WGAN has the lowest column-wise fidelity score, making the generated data less suitable for use cases that require precise replication of the original data. In contrast, although FinDiff has the lowest DCR score, it achieves a high NNDR score of 0.988 for the real dataset. This implies that even though FinDiff poses re-identification risks (low DCR), a high NNDR indicates that the nearest synthetic neighbors belong to the same clusters as the original records. There are notable deviations between the real and IBM simulated datasets, primarily characterized by lower NNDR scores for CTGAN, FinDiff, and TVAE in the IBM simulated dataset. This may suggest that their relative distances originate from regions with few data points, such as outliers. The simpler and sparse structure of the IBM simulated data may make it easier for some algorithms to replicate, potentially increasing the risk of tracing back to original records.

In conclusion, DGAN excels in scenarios where privacy is a priority, such as financial data sharing, and outperforms WGAN in terms of graph similarity. For tasks like upsampling or increasing data observations where data sharing is not a concern, FinDiff and TVAE are preferable due to their superior performance in data replication. Nonetheless, CTGAN offers a good balance across all metrics, making it suitable for general applications where a moderate level of privacy is acceptable. For the banking sector, this research offers valuable insights into selecting the most suitable generative models for privacy-sensitive data sharing, data augmentation, and achieving a balance between fidelity and efficiency. Future work should focus on improving these models' ability to replicate complex graph structures while enhancing privacy features to ensure compliance with regulatory standards.